\documentclass[10pt,twocolumn,letterpaper]{article}

\usepackage{iccv}
\usepackage{times}
\usepackage{epsfig}
\usepackage{graphicx}
\usepackage{amsmath}
\usepackage{amssymb}
\usepackage{booktabs}
\usepackage{tabularx}
\usepackage{bbm}
\usepackage[accsupp]{axessibility}
\usepackage[pagebackref=true,breaklinks=true,letterpaper=true,colorlinks,bookmarks=false]{hyperref}

\usepackage[breaklinks=true,bookmarks=false]{hyperref}

\iccvfinalcopy 

\def\iccvPaperID{6151} 

\ificcvfinal\fi

\begin{document}

\title{Leveraging SE(3) Equivariance for Learning 3D Geometric Shape Assembly}

\author{\textbf{Ruihai Wu}$^{1,3}$\thanks{Equal contribution, order determined by coin flip.} \quad  \textbf{Chenrui Tie}$^{2,1}$\footnotemark[1]  \quad  \textbf{Yushi Du}$^{2,1}$\footnotemark[1] \quad  \textbf{Yan Zhao}$^{1,3}$ \quad \textbf{Hao Dong}$ ^{1,3}$\thanks{Corresponding author}\\
$^1$CFCS, School of CS, PKU \quad 
$^2$School of EECS, PKU \quad 
\\$^3$National Key Laboratory for Multimedia Information Processing, School of CS, PKU \quad\\
{\tt\normalsize \{wuruihai,crtie,duyushi628,yan790,hao.dong\}@pku.edu.cn}
}

\maketitle
\ificcvfinal\thispagestyle{empty}\fi

\begin{abstract}
Shape assembly aims to reassemble parts (or fragments) into a complete object, which is 
a common task in our daily life.
Different from the semantic part assembly (\emph{e.g.}, assembling a chair's semantic parts like legs into a whole chair),
geometric part assembly (\emph{e.g.}, assembling bowl fragments into a complete bowl) is an emerging task in computer vision and robotics.
Instead of semantic information, this task focuses on geometric information of parts.
As the both geometric and pose space of fractured parts are exceptionally large, shape pose disentanglement of part representations is beneficial to geometric shape assembly.
In our paper, we propose to leverage SE(3) equivariance for such shape pose disentanglement.
Moreover, while previous works in vision and robotics only consider SE(3) equivariance for the representations of single objects,
we move a step forward and propose leveraging SE(3) equivariance for representations considering multi-part correlations,
which further boosts the performance of the multi-part assembly.
Experiments demonstrate the significance of SE(3) equivariance and our proposed method for geometric shape assembly.
Project page: \href{https://crtie.github.io/SE-3-part-assembly/}{https://crtie.github.io/SE-3-part-assembly/}

\end{abstract}

\section{Introduction}


Shape assembly aims to compose the parts or fragments of an object into a complete shape. 
It is a common task in the human-built world, from furniture assembly~\cite{lee2021ikea, zhan2020generative} (\emph{e.g.}, assemble chair parts like legs and handles into a whole chair) to fractured object reassembly~\cite{chen2022neural, sellan2022breaking} (\emph{e.g.}, assemble bowl fractures into a whole bowl) .
When trying to complete an object from parts, we will focus on their \textbf{\textit{geometric}} and \textbf{\textit{semantic}} information. 

There is a vast literature in both the computer vision and robotics fields studying the shape assembly problem, especially for the application purposes like furniture assembly and object assembly~\cite{agrawala2003designing,lee2021ikea,li2020learning,zhan2020generative}. 
Imagine we want to assemble a simple table with four wooden sticks and a flat board, we can infer that the sticks are the table legs so they should be vertically placed, while the board is the table top and should be horizontally placed. 
Here, we not only use geometric clues to infer the parts' functions but also use semantic information to predict the parts' poses. 

\begin{figure}[t]
  \centering
  \includegraphics[width=\linewidth, 
  ]{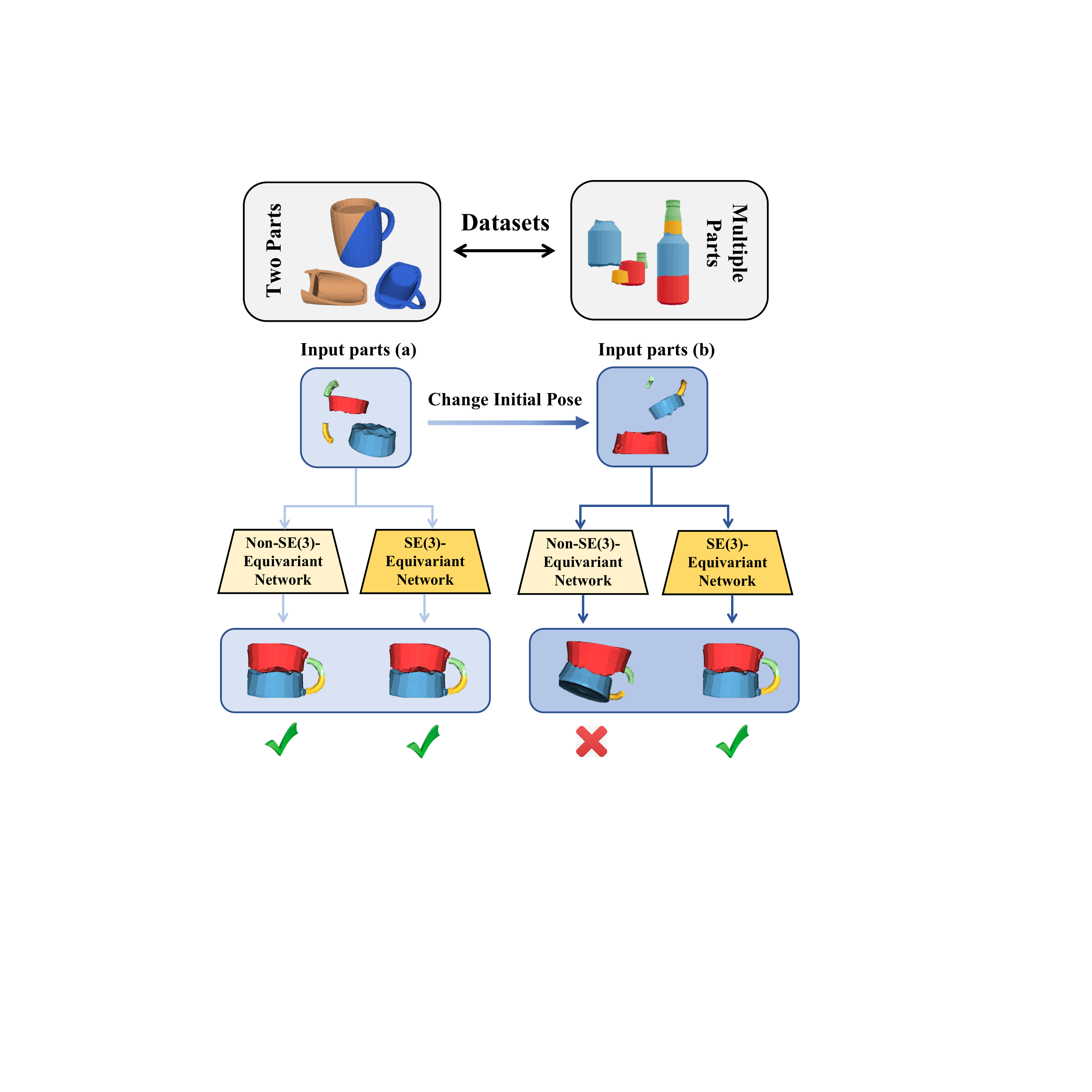}
  \caption{
   \textbf{Geometric Shape Assembly} aims to assemble different fractured parts into a whole shape. We propose to leverage \textbf{SE(3) Equivariance} for learning Geometric Shape Assembly, which disentangles poses and shapes of fractured parts, and performs better than networks without SE(3)-equivariant representations.
  }
  \label{fig_teaser}
\end{figure}

Recently, a two-part geometric mating dataset is proposed in NSM~\cite{chen2022neural}, which considers shape assembly from a pure geometric perspective,
without relying on semantic information. 
This work randomly cuts an object into two pairs, and studies how to mate the fragment pairs into the original shape. 
Such design is practical in some applications such as object kitting~\cite{devgon2021kit,li2022scene}, form fitting~\cite{zakka2020form2fit}, and protein binding~\cite{sverrisson2021fast}. 
In these tasks, the semantic information can hardly be acquired from the fragment shapes, and thus it is nearly impossible to predict fragments' poses relying on semantic information (e.g. part acts as a leg should be horizontally placed). 
Instead, such geometric mating tasks should be accomplished by relying on geometric cues.

Furthermore, the pairwise assembly task can be extended to the multi-part assembly task, and thus the pose space will grow much larger.
Recent work~\cite{sellan2022breaking} proposes a large-scale dataset named Breaking Bad, which models the destruction process of how an object breaks into fragments. For each object, there are multiple broken fragments with various and complex geometry, making it much more challenging for geometric shape understanding and assembly. Therefore, how to reduce the pose space and effectively assembly multiple fragments that are non-semantic but with diverse geometry still remains a problem.

Compared to furniture assembly, which relies on both part semantics and geometry,
geometric assembly that assembles diverse fractures mainly focuses on geometric information,
while the space of part pose and geometry are much larger in this task.
Therefore, \textbf{shape pose disentanglement} plays a significant role in boosting the performance of geometric shape assembly.

Recently, achieving SE(3) equivariance for object representations is arousing much attention in 3D computer vision and robotics.
Many works have studied SE(3)-equivariant architectures~\cite{chen2021equivariant,chen20223d,deng2021vector,fuchs2020se,katzir2022shape,thomas2018tensor,wang2019dynamic,weiler20183d,zhao2020quaternion} and leveraged SE(3) equivariance in object pose estimation~\cite{li2021leveraging, liu2023selfsupervised} or robotic object manipulation~\cite{kimse,ryu2022equivariant,simeonov2022seequivariant,simeonov2022neural,xue2022useek}.
SE(3) equivariance is suitable for the disentangling of shapes and poses of parts in geometric shape assembly.
Specifically, like previous works~\cite{chen2022neural, sellan2022breaking}, we formulate the shape assembly task as a pose prediction problem, and the target is to predict the canonical SE(3) pose for each given fragment to compose a whole shape.
For every single fragment, the predicted pose transformation should be \textit{equivariant} to its original pose, while being \textit{invariant} to other fragments' poses.
Accordingly, the learned representations have two main features: \textit{consistency} and \textit{stability}.
\textit{Consistency} means that parts with the same geometry but different poses should have \textit{equivariant} representations, while \textit{stability} means the representation of a specific part should be \textit{invariant} to all other parts' poses and only related to their geometry characteristics. 
Leveraging such properties,
the network can reduce the large pose space of the complex geometric shape assembly task and thus focus on the fragments' geometric information for shape assembly.

While most previous works in vision and robotics only leverage SE(3) equivariance representations on a single shape,
there exist multiple complex fractured parts in our geometric shape assembly task,
and 
extracting 
other parts' geometric information is essential to a successful reassembly.
How to leverage
SE(3)-equivariant representations for multi-parts shape assembly is not a trivial problem, 
as learned part representations should not only consider the certain part, but also consider correlations with other parts (\emph{e.g.}, whether the notches of two parts match each other),
while keeping the equivariance property.
We propose to utilize both equivariant and invariant representations of single parts to compose the equivariant part representations including part correlations.
To the best of our knowledge,
we are the first to leverage the SE(3) equivariance property among multiple objects.

In summary, we make the following contributions:
\begin{itemize}
    \item We propose to leverage SE(3) equivariance that disentangles shapes and poses of fractured parts for geometric shape assembly.
    \item Utilizing both SE(3)-equivariant and -invariant representations, we learn SE(3)-equivariant part representations with part correlations for multi-part assembly.
    \item Experiments on representative benchmarks, including both two-part and multi-part 3D geometric shape assembly, demonstrate the superiority of SE(3) equivariance and our proposed method.
\end{itemize}











\section{Related Works}
\subsection{3D Shape Assembly}
Shape assembly is a long-standing problem with a rich literature. Many works have been investigating how to construct a complete shape from given parts~\cite{chen2022neural,funkhouser2011learning,jones2021automate,lee2021ikea,li2020learning,narayan2022rgl,willis2022joinable,yin2020coalesce,zhan2020generative}, especially in application-specific domains. 
Based on PartNet, a large-scale dataset that contains diverse 3D objects with fine-grained part information, previous works propose a dynamic graph learning method~\cite{zhan2020generative} to predict 6-DoF poses for each input part (\emph{e.g.}, the back, legs and bars of a chair) and then assemble them into a single shape as output, or study how to assemble 3D shape given a single image depicting the complete shape~\cite{li2020learning}. 
Besides, many works study the shape assembly problem for different applications like furniture assembly~\cite{lee2021ikea},
or unique needs of CAD workflow~\cite{jones2021automate}.

However, most previous works rely deeply on the
semantic information of object parts, sometimes bypassing the geometric cues. 
As for the geometric cues, a recent work, NSM~\cite{chen2022neural}, tries to solve the two-part mating problem by mainly focusing on shape geometries without particular semantic information. Besides, a new dataset, Breaking Bad~\cite{sellan2022breaking}, raises a new challenge about how to assemble multiple non-semantic fragments into a complete shape. This work demonstrates that fractured shape reassembly is still a quite open problem. Following these two works, we focus on studying the geometric information and tackling the pure geometric shape assembly problem.

\subsection{SE(3)-Equivariant Representations}
Recently, achieving SE(3) equivariance has attracted a lot of attention, and many SE(3)-equivariant architectures have emerged~\cite{chen2021equivariant,chen20223d,fuchs2020se,katzir2022shape,thomas2018tensor,wang2019dynamic,weiler20183d,zhao2020quaternion}. 
Thomas \emph{et al.}~\cite{thomas2018tensor} propose a tensor field neural network that uses filters built from spherical, and Deng \emph{et al.}~\cite{deng2021vector} introduce Vector Neurons that can facilitate rotation equivariant neural networks by lifting standard neural network representations to 3D space. We follow Vector Neuron~\cite{deng2021vector} and apply the vector neuron version of DGCNN~\cite{wang2019dynamic} model in our pipeline. 

Meanwhile, many recent works have utilized equivariant models for point cloud registration~\cite{lin2022coarse}, object detection~\cite{yu2022rotationally}, pose estimation~\cite{li2021leveraging,liu2023selfsupervised}, robotic manipulation~\cite{kimse,ryu2022equivariant,simeonov2022seequivariant,simeonov2022neural,xue2022useek}, and demonstrate that such equivariant models can significantly improve the sample efficiency and generalization ability.
In this paper, we leverage SE(3)-equivariant representations for geometric shape assembly to disentangle the shape and the pose.






\section{Problem Formulation}
\label{sec:formulation}

Imagine an object has been broken into $N$ fractured parts (\emph{e.g.}, a broken porcelain vase found during archaeological work), we obtain the point cloud of each part, which forms $\mathcal{P} = \{P_i\}_{i=1}^N$. Our goal is to assemble these parts together and recover the complete object.

Formally, our framework takes all parts' point cloud $\mathcal{P}$ as the input and predicts the canonical 3D pose of each part. 
We denote the predicted SE(3) pose of the $i$-th fractured part as $(R_i,\ T_i)$,
where $R_i \in \mathbb{R}^{3 \times 3}$ is the predicted rotation matrix and $T_i \in \mathbb{R}^{3}$ is the predicted translation vector. 
Then, we apply the predicted pose to transform the point cloud of each part and get the $i$-th part's predicted point cloud $P'_i =  P_i R_i + T_i$. 
The union of all the transformed point clouds $P'_{whole} = \bigcup_i P'_i$ is our predicted assembly result.
\section{Method}

\begin{figure*}[!]
  \centering
\includegraphics[width=\linewidth, 
]{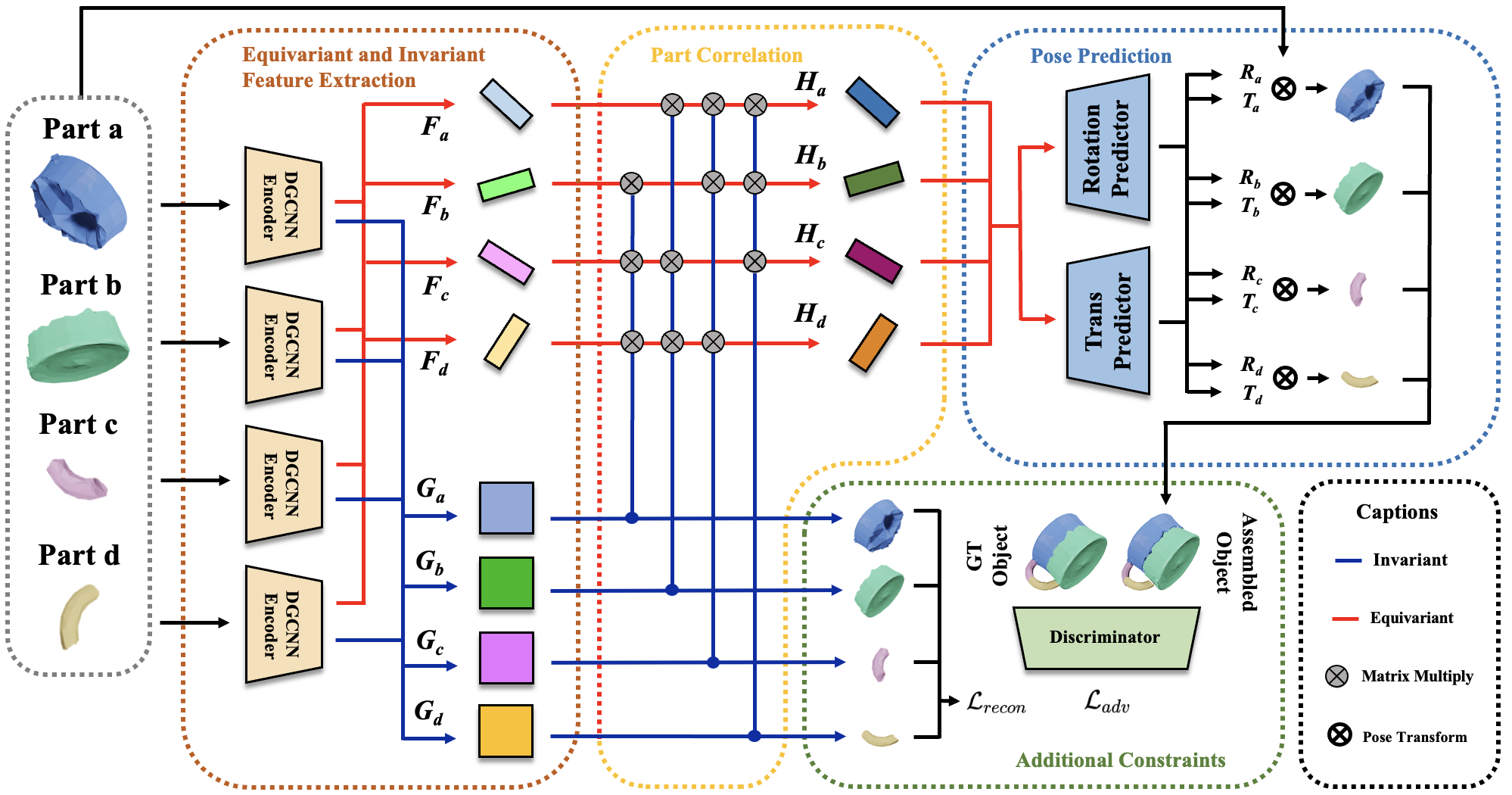}
  \vspace{0.5mm}
  \caption{\textbf{Overview of our proposed framework.}
  Taking as input the point cloud of each part $i$,
  our framework first outputs the equivariant representation $F_i$ and invariant representation $G_i$,
  computes the correlation between part $i$ and each part $j$ using the matrix multiplication of $F_i$ and $G_j$, and thus gets each part's equivariant representation $H_i$ with part correlations.
  The rotation decoder and the translation decoder respectively take $H$ and decode the rotation and translation of each part.
  Additional constraints such as adversarial training and canonical point cloud reconstruction using $G$ further improves the performance of our method.
  }
  \label{fig_framework}
\end{figure*}

Our method leverages SE(3)-equivariant representations 
for geometric shape assembly.
We start the Method Section by describing how to leverage SE(3)-equivariant for a single part as a basis (Sec.~\ref{sec:se3_part_repr}). 
Then, as geometric shape assembly requires each part to consider its correlations with other parts,
we describe extending the leverage of SE(3) equivariance from single-part representations to part representations considering correlations with other parts (Sec.~\ref{sec:se3_multi_part_repr}).
Based on the learned equivariant representations and apart from predicting the pose of each fractured part,
to further ensure all the re-posed parts compose a whole object, we propose translation embedding (Sec.~\ref{sec:trans}) for geometric assembly and use adversarial learning (Sec.~\ref{sec:adv}).
Finally, we describe the loss functions (Sec.~\ref{sec:loss}).



\subsection{Leveraging SE(3) Equivariance for Single Parts}
\label{sec:se3_part_repr}

For the brevity of description, we first introduce a simple version (as a basis of our whole method): leveraging SE(3) equivariance in single parts' representations, without considering the correlations between multiple parts.

Specifically, in this section, we start by revisiting Vector Neurons Networks (VNN)~\cite{deng2021vector}, a general framework for SO(3)-equivariant (rotation equivariant) network. Leveraging VNN, we introduce how we leverage rotation equivariance and translation equivariance for single parts.


\paragraph{Vector Neurons Networks (VNN)}
is a general framework for SO(3)-equivariant networks. 
It extends neurons from 1D scalars to 3D vectors and provides various SO(3)-equivariant 
neural operations including
linear layers (such as Conv and MLP), non-linear layers (such as Pooling and ReLU) and normalization layers. 
Besides, it also designs SO(3)-invariant layers to extract SO(3)-invariant representations.
The above properties are mathematically rigorous.


\paragraph{Rotation Equivariance and Invariance.}

In geometric part assembly, we suppose the predicted rotation of a part is equivariant with its original orientation and invariant with other parts' orientation.
Accordingly, the network should learn both equivariant and invariant representations for each part.
Based on VNN, we build a DGCNN~\cite{wang2019dynamic} encoder with a SO(3)-equivariant head $\mathcal{E}_{equiv}$ and a SO(3)-invariant encoder head $\mathcal{E}_{inv}$ to extract part features with corresponding properties.
Specifically, given an input point cloud $P$, and a random rotation matrix $R$, the encoders $\mathcal{E}_{equiv}$ and $\mathcal{E}_{inv}$ respectively satisfy rotation equivariance and invariance:
\begin{equation}
\mathcal{E}_{equiv}(PR)=\mathcal{E}_{equiv}(P)R
\end{equation}

\begin{equation}
    \mathcal{E}_{inv}(PR)=\mathcal{E}_{inv}(P)
\end{equation}

\paragraph{Translation Equivariance.}
\label{sec:trans_eqv}
To achieve translation equivariance in parts' pose prediction, we preprocess the raw point cloud of each part by posing its gravity center on the coordinate origin.
That's to say, with an input point cloud $P=(p_1,p_2,...,p_n), p_i\in \mathbb{{R}}^3$, where $n$ is the number of points, 
we compute its gravity center $\hat{x}=(\sum_{i=1}^np_i)/n$, and get the preprocessed point cloud $\tilde{P}=P-\hat{x}$,
and then we use $\tilde{P}$ as the network input. 
In this way, our prediction is translation equivariant.
Formally, let $T_{pred}$ denote the predicted translation output,
if the part's point cloud changes from $P$ to $P + \Delta T$, we have:
\begin{equation}
\begin{split}
    T_{pred}(P+\Delta T) = T_{pred}(P)+\Delta T
\end{split}
\end{equation}


\subsection{Leveraging SE(3) Equivariance for Parts with Part Correlations}
\label{sec:se3_multi_part_repr}


In the geometric shape assembly task, all the fractured parts should be reassembled together, with each edge matching up well with the others. Therefore, all parts' geometry information, especially the geometry of edges, is significant.
Besides, it is necessary to analyze all parts' shape information together to infer the position of each individual part, otherwise, it would be nearly impossible to predict accurate positions for those parts. Therefore, the correlations between parts are essential in the geometric assembly task, and we propose a correlation module to aggregate the information of multiple parts, while keeping leveraging SE(3) equivariance.

Note that the translation equivariance in part pose predictions can be achieved using the same approach in Sec.~\ref{sec:trans_eqv}, so in this section we mainly describe how to leverage rotation equivariance when taking multiple parts.

\paragraph{Rotation Equivariance in Multi-Part Representations.}
To predict the final pose of part $i$, we should consider its correlations with other parts.
In fact, what really matters for this part's pose prediction is other parts' shape geometry instead of their initial poses.
In other words, changed initial poses of other parts may not affect the predicted pose of part $i$.
Therefore, it comes naturally that we leverage the rotation-invariant representations of other parts (which are invariant to their initial poses) to extract their geometric features and further compute their correlations with part $i$.

Specifically, given the point cloud $P_i\in \mathbb{R}^{n\times3}$ of the $i$-th part, 
we pass it through rotation-equivariant and -invariant encoders $\mathcal{E}_{equiv}$ and $\mathcal{E}_{inv}$ and get corresponding features (shown in the \textbf{Equivariant and Invariant Feature Extraction} module in Figure~\ref{fig_framework}):
\begin{equation}
\begin{split}
F_i=\mathcal{E}_{equiv}(P_i), \quad F_i\in\mathbb{R}^{f\times3} \\
G_i=\mathcal{E}_{inv}(P_i), \quad G_i\in\mathbb{R}^{f\times f}
\end{split}
\end{equation}

As shown in the \textbf{Part Correlation} module in Figure~\ref{fig_framework},
to extract the correlation feature $C_{i,\,j}$ between part $i$ and part $j,(j \ne i)$, we use matrix multiplication between $G_j$ and $F_i$:

\begin{equation}
C_{i,\,j}=G_j\cdot F_i, \quad C_{i,j}\in\mathbb{R}^{f\times 3}
\label{equ:Cij}
\end{equation}
where $\cdot$ denotes \textbf{matrix multiplication}.

As $G_j$ is invariant to $P_j$, and $F_i$ is equivariant to $P_i$, the matrix multiplication $C_{i,\,j}$ is thus equivariant to $P_i$ with the geometry correlation between $P_i$ and $P_j$.

Furthermore, to get the part representation $H_i$ considering correlations with all other parts while maintaining equivariance with $P_i$,
we define $H_i$ as:

\begin{equation}
H_i=\frac{1}{N-1}\sum_{j=1,\,j\ne i}^{N}G_j\cdot F_i, \quad H_{i}\in\mathbb{R}^{f\times 3}
\label{equ:H}
\end{equation}

As $G_j$ is invariant with $P_i$ and $F_i$ is equivariant with $P_i$, it's easy to verify that, $H_i$ is equivariant with $P_i$, and invariant with $P_j(j\ne i)$.  \emph{i.e.}, for any rotation matrix $R$ applied on part $i$ or any other part $j$:
\begin{equation}
\begin{split}
&H_i(P_1,\,...,\,P_iR,\,...P_N)=H_i(P_1,\,...,\,P_i,\,...,\,P_N)R\\
&H_i(P_1,\,...,\,P_jR,\,...P_N)=H_i(P_1,\,...,\,P_j,\,...,\,P_N), (j\ne i)
\label{requirements}
\end{split}
\end{equation}



\paragraph{Pose Prediction.}
As shown in the \textbf{Pose Prediction} module in Figure~\ref{fig_framework}, given the equivariant representation $H_i$ of part $i$ with part correlations,
we use a pose regressor $\mathcal{R}$ to predict its rotation $R_{pred,\ i}$ and translation $T_{pred,\ i}$:
\begin{equation}
\begin{split}
    R_{pred,\ i},\ T_{pred,\ i} = \mathcal{R}(H_i), \\ R_{pred,\ i}\in\mathbb{R}^{3\times 3}, \quad T_{pred,\ i}\in\mathbb{R}^{3}
\end{split}
\end{equation}

\paragraph{Canonical Part Reconstruction.}
\label{sec:recon}
To ensure that the rotation invariant feature $G_i$ encodes geometric information of $P_i$ with any initial pose, we use a point cloud decoder $\mathcal{D}$ and expect $\mathcal{D}$ to decode point cloud of $i$-th part in the canonical view when receiving $G_i$ (shown in the \textbf{Additional Constraint} module in Figure~\ref{fig_framework}):
\begin{equation}
 P^*_{pred,\ i} = \mathcal{D}(G_i), \quad  P^*_{pred,\ i}\in\mathbb{R}^{n\times 3}
\label{equ:decoder}
\end{equation}
Let $P^*_{gt,i}$ denote the canonical point cloud of $P_i$ and $P^*_{pred,\,i}$ denote the prediction,
we minimize the Chamfer Distance between $P^*_{pred,\,i}$ and $P^*_{gt,\,i}$.

\subsection{Translation Embeddings for Part Representations}
\label{sec:trans}

Since the reassembled whole object is the composition of multiple re-posed parts,
although the above described designs learn the pose of each part,
the framework lacks leveraging the property that the representations of all parts could compose the whole object.

Inspired by Visual Translation Embedding (VTransE) \cite{hung2020contextual,zhang2017visual} that maps different objects' features into a space where the relations between objects can be the feature translation,
we propose a similar Translation Embedding where the representations of parts can be added up to the representation of the whole shape.

Formally, denoting the point cloud of the whole object at canonical pose as $P^*_{gt}$, we pass it through our rotation equivariant encoder to get $F^*_{gt}=\mathcal{E}_{equiv}(P^*_{gt})$, and minimize:
\begin{equation}
    \mathcal{L}2(\sum_iH_i,\ F^*_{gt})
\end{equation}
where $H_i$ is the rotation equivariant feature of $i$-th fracture.

Through this procedure, rotation equivariant representations of parts would be more interpretable as a whole.

\subsection{Adversarial Learning}
\label{sec:adv}
The above described designs use proposed representations to learn the pose of each part,
lacking the evaluation that all re-posed parts visually make up a whole shape.
Following the design of~\cite{chen2022neural},
we employ a discriminator $\mathcal{M}$ and use adversarial learning to make the re-posed parts visually look like those of a whole object, as shown in the \textbf{Additional Constraint} module in Figure~\ref{fig_framework}.

Our discriminator $\mathcal{M}$ takes as input the predicted reassembly shape $P'_{whole}$ (defined in Sec.~\ref{sec:formulation}) and the ground truth point cloud of the whole object $P_{whole}$,
and distinguishes whether the input point clouds look visually plausible like a complete object.
To achieve this, we define a loss term $\mathcal{L}_\text{G}$ for training the generator ($i.e.$ encoders $\mathcal{E}$ and pose regressor $\mathcal{R}$), which is defined as:
\begin{equation}
  \mathcal{L}_{G} = \mathbb{E}_{}\big[\|\mathcal{M}(P'_{whole}) - 1\|\big],
  \label{eq:G}
\end{equation}
and an adversarial loss $\mathcal{L}_{D}$ for training the discriminator $\mathcal{M}$, which is defined as:
\begin{equation}
  \mathcal{L}_{D} = \mathbb{E}_{}\big[\|\mathcal{M}(P'_{whole})\|\big] + \mathbb{E}_{}\big[\|\mathcal{M}(P_{whole}) - 1\|\big]
  \label{eq:adv}
\end{equation}

Through the adversarial training procedure, the reassembled shapes become more plausible as a whole.

\subsection{Losses}
\label{sec:loss}
Our loss function consists of the following terms:
\begin{equation}
\begin{split}
    \mathcal{L}=\lambda_{rot} \mathcal{L}_{rot}+\lambda_{trans} \mathcal{L}_{trans}+\lambda_{point} \mathcal{L}_{point}\\ +\lambda_{recon}\mathcal{L}_{recon}+
    \lambda_{embed}\mathcal{L}_{embed}+\lambda_{adv} \mathcal{L}_{adv}
\end{split}
\end{equation}

For an input broken object, we sample point clouds from every fractured part and form $\mathcal{P} = \{P_i\}_{i=1}^N$. 
For the $i$-th part, we denote its ground truth rotation matrix and translation as $R_{gt,\ i}$ and $T_{gt,\ i}$,
and the predicted rotation matrix and translation as $R_{pred,\ i}$ and $T_{pred,\ i}$.

For rotation, we use geodesic distance (GD) between $R_{gt}$ and $R_{pred}$ as our rotation loss:
\begin{equation}
   \mathcal{L}_{rot} =arccos\frac{tr(R_{gt}R_{pred}^T)-1}{2}
\end{equation}

For translation, we use $\mathcal{L}2$ loss between $T_{gt}$ and $T_{pred}$ as the translation prediction loss:
\begin{equation}
    \mathcal{L}_{trans} = \mathcal{L}2(T_{pred},\ T_{gt})
\end{equation}

Following~\cite{sellan2022breaking}, we use Chamfer Distance to further jointly supervise the predicted translation and rotation by supervising the predicted re-posed point cloud:
\begin{equation}
    \mathcal{L}_{point} = Chamfer(PR_{pred}+T_{pred},\ PR_{gt}+T_{gt})
\end{equation}

As mentioned in Sec.~\ref{sec:recon}, we also use Chamfer Distance as the reconstruction loss to supervise the invariant representation $G_i$ can be further decoded to a canonical point cloud $P^*_{pred,i}$, ensuring $G_i$ encodes geometric information with any initial pose:
\begin{equation}
    \mathcal{L}_{recon} = Chamfer(P^*_{pred,i},\ P_iR_{gt,i}+T_{gt,i})
\end{equation}

From Sec.~\ref{sec:trans}, we design translation embedding loss to supervise that the representations of all fractured parts can be added up to the representation of the complete shape: 
\begin{equation}
    \mathcal{L}_{embed} = \mathcal{L}2(\sum_iH_i,\ F^*_{gt})
\end{equation}

From Sec.~\ref{sec:adv}, the adversarial loss is defined as:
\begin{equation}
    \mathcal{L}_{adv} = \mathbbm{1}_D\mathcal{L}_{D}+\mathbbm{1}_G\mathcal{L}_{G}
\end{equation}
where $\mathbbm{1}_D =1 $ only if we're updating discriminator,  $\mathbbm{1}_G =1 $ only if updating generator.

\section{Experiments}

\subsection{Datasets, Settings and Metrics}

\paragraph{Datasets.} We use two benchmark datasets for evaluation: 
\begin{itemize}
  \item Geometric Shape Mating dataset~\cite{chen2022neural} for \textbf{two-part assembly (mating)}. Objects in this dataset are cut into two parts by the randomly generated heightfields that can be parameterized by different functions. Specifically, we employ 4 kinds of cut types (planar, sine, parabolic and square functions) on 5 categories of objects (Bag, Bowl, Jar, Mug and Sofa) in ShapeNet~\cite{shapenet2015}.
  We employ the official data collection code, collect 41,000 cuts for training and 3,100 cuts for testing.
  \item Breaking Bad dataset's commonly used ``everyday'' object subset~\cite{sellan2022breaking} 
for \textbf{multi-part assembly}.
Compared with Geometric Shape Mating dataset, this dataset is much more challenging, as the objects are irregularly broken into multiple fragments by physical plausible decomposition , resulting in more parts with much more complex geometries.
Our study focuses more on this multi-part geometric assembly problem.
\end{itemize} 

On both datasets, we train all methods in all categories, and test them on unseen objects in the same categories.

\paragraph{Metrics.} Following the evaluation metrics of the two datasets~\cite{chen2022neural, sellan2022breaking}, 
we import geodesic distance (GD) to measure the difference between predicted rotation and ground truth rotation.
To further evaluate both the rotation and translation prediction,
we compute the root mean squared error RMSE ($R$) between the predicted rotation $R$ and the corresponding ground truth values, and the root mean squared error RMSE ($T$) between the predicted translation $T$ and the corresponding ground truth values.
Here we use Euler angle to represent rotation.
 
Besides, we follow the evaluation protocol in~\cite{li2020learning, sellan2022breaking} and adopt part accuracy (PA) as an evaluation metric. This metric measures the portion of `correctly placed' parts. We first use predicted rotation and translation to transform the input point cloud, and then compute the Chamfer Distance between the transformed point cloud and the ground truth. If the distance is smaller than a threshold, we count this part as `correctly placed'.
\paragraph{Hyper-parameters.} We set batch size to be 32 for Breaking Bad, 48 for Geometric Shape Mating, and the initial learning rate of Adam Optimizer~\cite{kingma2014adam} to be 0.0001. We train the model 80 and 120 epochs respectively for Geometric Shape Mating and Breaking Bad.

\subsection{Baselines}

For the Geometric Shape Mating dataset, the two-part geometric shape assembly task, we compare our method with NSM~\cite{chen2022neural}, the state-of-the-art method for two-part mating.
For the Breaking Bad dataset, the multi-part geometric shape assembly task, 
we modified the official code of the NSM~\cite{chen2022neural} from two-part geometric shape assembly to multi-part geometric assembly by predicting the pose of each input part. We also compare our method with DGL~\cite{zhan2020generative} and LSTM~\cite{hochreiter1997long} following the Breaking Bad benchmark~\cite{sellan2022breaking}.
All baseline implementations use the official code in two benchmarks~\cite{chen2022neural, sellan2022breaking}.
The baselines are described as follows:

\begin{itemize}
  \item 
  \textbf{NSM}~\cite{chen2022neural} extracts part features using transformer and predicts their poses for mating, achieving state-of-the-art performance in two-part mating.
  \item 
  \textbf{DGL}~\cite{sellan2022breaking, zhan2020generative} uses graph neural networks to encode and aggregate part features, and predicts the pose of each part. Following~\cite{sellan2022breaking}, we remove the node aggregation procedure as there does not exist parts with the same geometric appearance in geometric assembly.
  \item 
  \textbf{LSTM}~\cite{sellan2022breaking, zhan2020generative, hochreiter1997long} uses bi-directional LSTM to take part features as input and sequentially predicts the pose of each part. This method assembles the decision-making method of humans when faced with geometric shape assembly problems.
\end{itemize} 


\begin{figure*}[h]
  \centering
  \includegraphics[width=\linewidth,
  ] {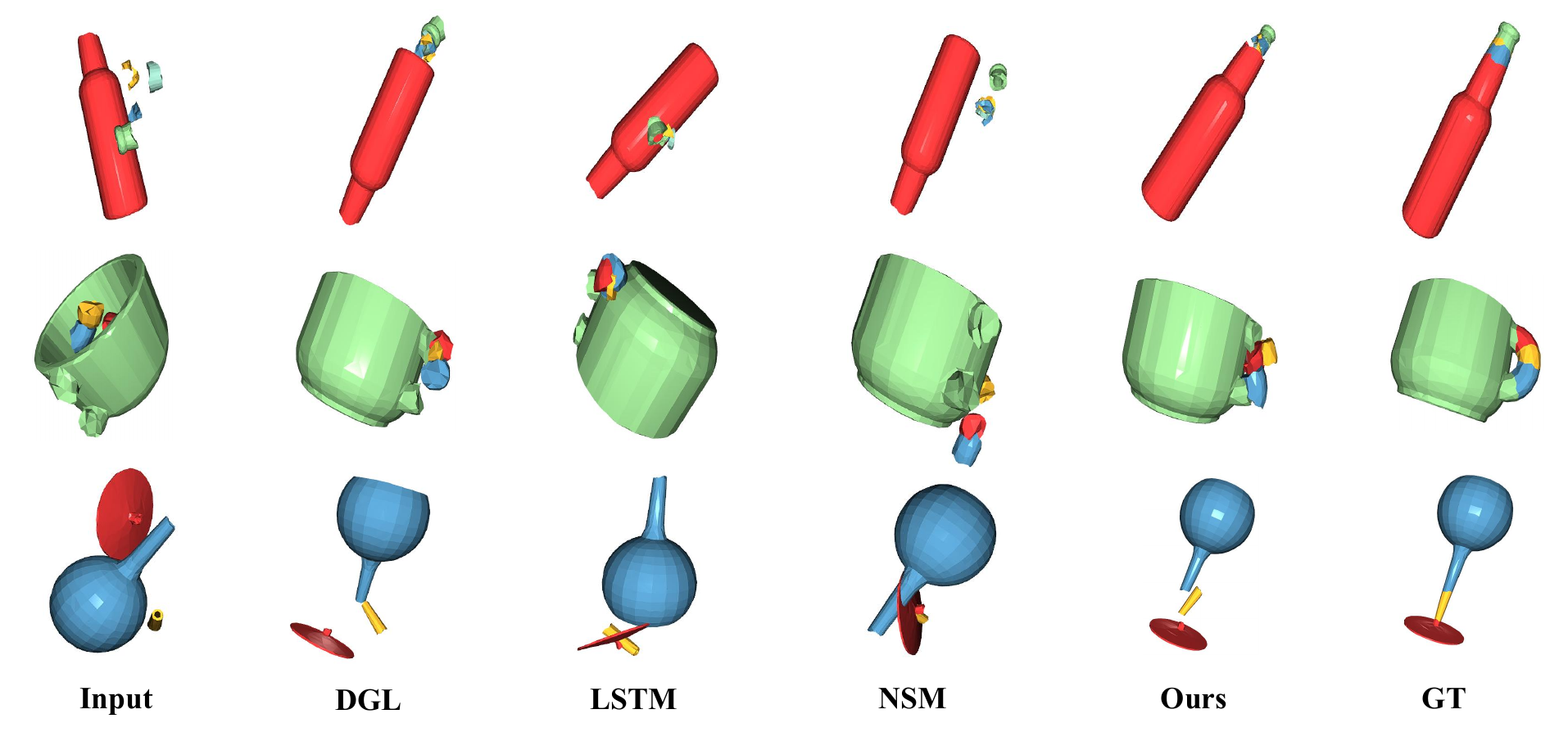}
  \caption{\textbf{Qualitative results on Breaking Bad dataset for multi-part geometry shape assembly.
  }
  We observe better rotation and translation predictions (especially rotation) than baseline methods.
  }
  \vspace{0mm}
  \label{fig_visu_breaking}
\end{figure*}

\begin{figure}[h]
  \centering
  \includegraphics[width=\linewidth, trim={1cm, 0.4cm, 1cm, 0.2cm}, clip
  ] {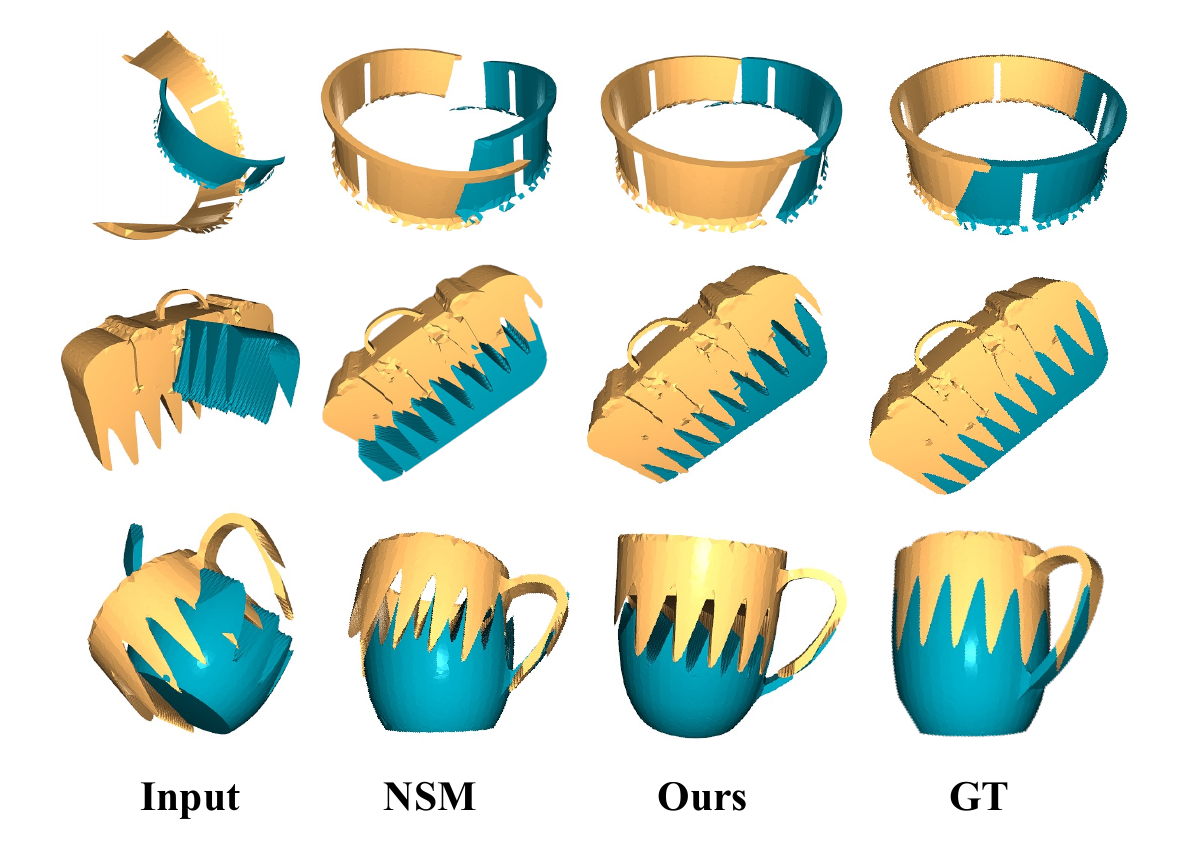}
  \caption{\textbf{Qualitative results on Geometric Shape Mating dataset for two-part geometric shape assembly.} 
  We observe better pose predictions (especially rotation) than NSM.
  }
  \label{fig_visu_nsm}
\end{figure}

\begin{table}[h]
  \begin{center}
  \small
  {
  \begin{tabular}{lcccccc}
    \toprule
    Method & RMSE ($R$) $\downarrow$ & GD ($R$) $\downarrow$ & RMSE ($T$) $\downarrow$  &  PA $\uparrow$ \\ 
    \midrule
    & degree & rad & $\times 10^{-2}$ &  $\%$ \\  
    \midrule
    DGL & 84.1 & 2.21 & 14.7 & 22.9\\ 
    LSTM & 87.6 & 2.24 & 16.3 & 13.4\\ 
    NSM & 85.6 & 2.21 & 15.7 & 16.0\\
    Ours & $\textbf{75.3}$ & $\textbf{2.00}$ & $\textbf{14.1}$ & $\textbf{26.7}$ \\ 
    \bottomrule
  \end{tabular}
  }
\vspace{.2cm}
  \caption{
  \textbf{Quantitative evaluation on Breaking Bad dataset for multi-part geometric assembly.}
  We report quantitative results of our method and three learning-based shape assembly baselines on the \texttt{everyday} object subset. }
  \label{table:bb_compare}
  \end{center}
\end{table}

\begin{table}[h]
  \begin{center}
  \small
  
  {
  \begin{tabular}{lcccccc}
    \toprule
    Method & RMSE ($R$) $\downarrow$ & GD ($R$) $\downarrow$ & RMSE ($T$) $\downarrow$  &  PA $\uparrow$ \\ 
    \midrule
    & degree & rad & $\times 10^{-2}$ &  $\%$ \\  
    \midrule
    NSM & 21.3 & 0.52 & 2.9 & 79.1\\
    Ours & $\textbf{15.9}$ & $\textbf{0.39}$ & $\textbf{2.7}$ & $\textbf{85.7}$ \\ 
    \bottomrule
  \end{tabular}
  }
  \vspace{.2cm}
\caption{
  \textbf{Quantitative evaluation on Geometric Shape Mating dataset for two-part geometric assembly.}
  We report quantitative results of our method and the NSM baseline.
  }
  \label{table:nsm}
  \end{center}
  \vspace{-7mm}
\end{table}

\subsection{Experimental Results and Analysis}
Table \ref{table:bb_compare} and \ref{table:nsm} show the quantitative performance of our method and baselines. 
The experimental results demonstrate that our method performs better than all baselines in both two-part and multi-part geometric shape assembly tasks over all evaluation metrics.

As discussed in~\cite{sellan2022breaking}, 
predicting rotations of multiple parts is pretty more difficult than translation. 
Table~\ref{table:bb_compare} and~\ref{table:nsm} show our method has a significant improvement in this aspect, and outperforms all baselines in the root mean squared error RMSE ($R$) metric and the geodesic distance (GD) metric.
In particular, our rotation is around 10 degrees less than the baselines. 
For translation prediction, our RMSE ($T$) also outperforms all baselines on both datasets.
In addition, our method also outperforms all baselines in part accuracy (PA), especially for the LSTM and NSM in the more challenging Breaking Bad dataset.

This may result from our SO(3) equivariant network that disentangles shape and pose information, reducing the difficulty of learning rotations of different posed parts with different geometries, thus allowing for better predictions.
%

Figure~\ref{fig_visu_nsm} shows qualitative examples of our method and NSM on the Geometric Shape Mating dataset.
Although it is a comparatively simple dataset and the task is nearly solved by previous methods, our method still performs better, especially in rotation prediction.

Figure~\ref{fig_visu_breaking} shows qualitative comparisons between our method and baselines on the more challenging and realistic Breaking Bad dataset.
Although this task is highly difficult and all methods could not solve the task, our method can better predict the pose (especially the rotation) of each part.

\begin{figure}[h]
  \centering
  \includegraphics[width=\linewidth, trim={3cm, 4cm, 4cm, 1.2cm}, clip
  ] {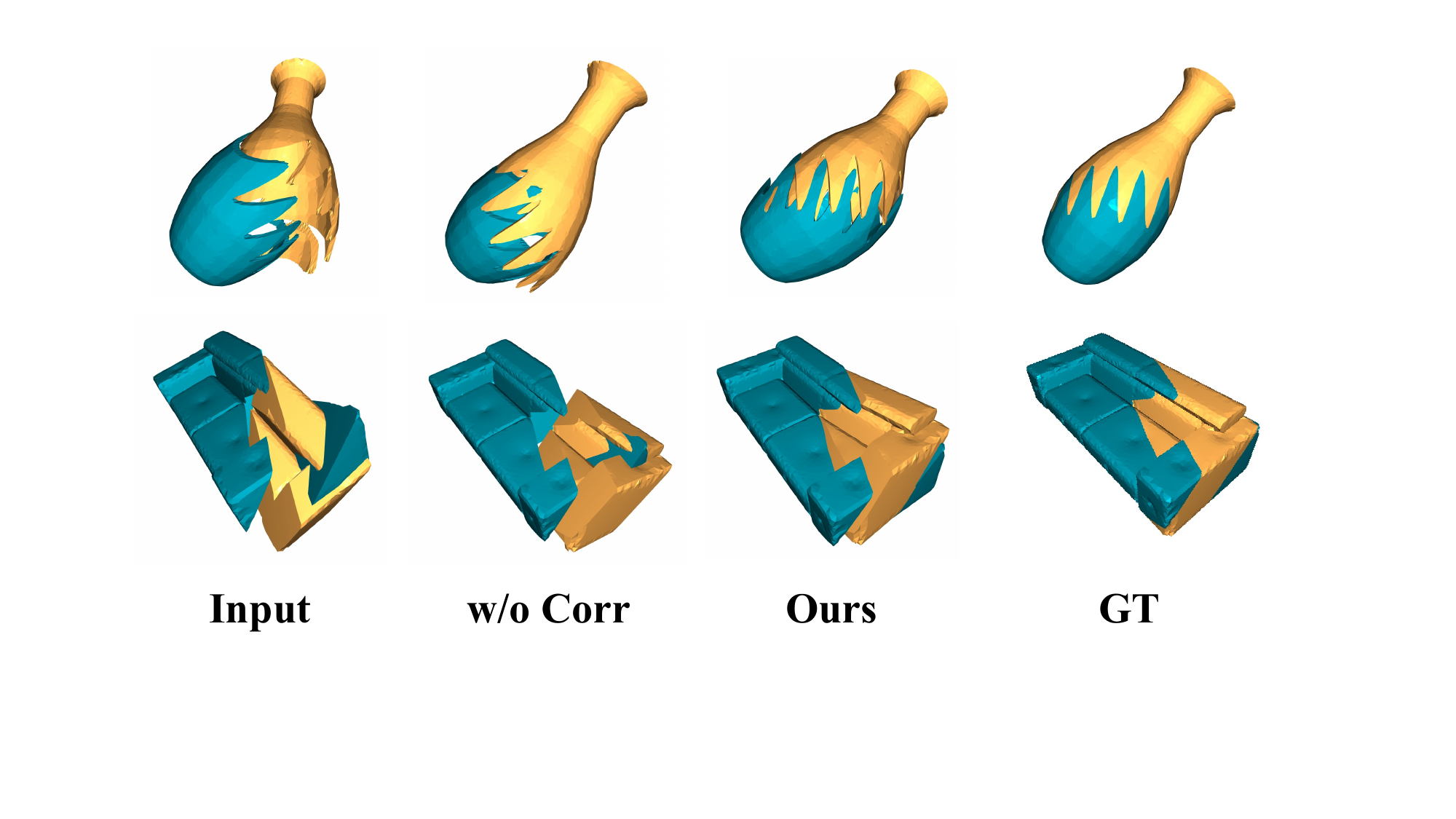}
  \caption{\textbf{Qualitative results of our method with and without part correlation on Geometric Shape Mating dataset.} 
  The parts with representations considering part correlations match better.
  }
  \vspace{0mm}
  \label{fig_visu_ab_corr}
\end{figure}


\subsection{Ablation Studies}

To further evaluate the effectiveness of different components in our framework,
we conduct ablation studies by comparing our method with the following ablated versions:

\begin{itemize}
  \item 
  \textbf{w/o Corr}: our method without considering part correlations in each part's equivariant representations.
  \item 
  \textbf{w/o TE}: our method without translation embedding.
  \item 
  \textbf{w/o Adv}: our method without adversarial learning.
\end{itemize}

\begin{table}[h]
  \begin{center}
  \small
  
  {
  \begin{tabular}{lcccccc}
    \toprule
    Method & RMSE ($R$) $\downarrow$ & GD ($R$) $\downarrow$ & RMSE ($T$) $\downarrow$  &  PA $\uparrow$ \\ 
    \midrule
    & degree & rad & $\times 10^{-2}$ &  $\%$ \\  
    \midrule
    w/o Corr & 19.2 & 0.52 & 2.9 & 80.5\\ 
    w/o TE & 17.3 & 0.46 & 2.8 & 84.3 \\ 
    w/o Adv & 16.7 & 0.43 & 2.8 & 82.6\\

    Ours & $\textbf{15.9}$ & $\textbf{0.39}$ & $\textbf{2.7}$ & $\textbf{85.7}$ \\ 
    \bottomrule
  \end{tabular}
  }
    \vspace{.2cm}
\caption{
  \textbf{Ablations on Geometric Shape Mating.} We compare with versions removing part correlations (w/o Corr), translation embedding (w/o TE) and adversarial learning (w/o Adv).
  }
  \label{table:NSMab}
  \end{center}
\end{table}
\vspace{-5mm}

\begin{table}[h]
  \begin{center}
  \small
  {
  \begin{tabular}{lcccccc}
    \toprule
    Method & RMSE ($R$) $\downarrow$ & GD ($R$) $\downarrow$ & RMSE ($T$) $\downarrow$  &  PA $\uparrow$ \\ 
    \midrule
    & degree & rad & $\times 10^{-2}$ &  $\%$ \\  
    \midrule
    w/o Corr & 79.8 & 2.17 & 15.7 & 18.4\\
    w/o TE & 77.2 & 2.04 & 15.2 & 22.5 \\
    w/o Adv & 77.6 & 2.02 & 14.3 & 23.7 \\ 
    Ours & $\textbf{75.3}$ & $\textbf{2.00}$ & $\textbf{14.1}$ & $\textbf{26.7}$ \\ 
    \bottomrule
  \end{tabular}
  }
    \vspace{.2cm}
    \caption{
  \textbf{Ablations on Breaking Bad.} We compare with versions removing part correlations (w/o Corr), translation embedding (w/o TE) and adversarial learning (w/o Adv). 
  }
  
  \label{table:BBab}
  \end{center}
\vspace{-5mm}
\end{table}

As shown in Table~\ref{table:NSMab} and~\ref{table:BBab}, and Figure~\ref{fig_visu_ab_corr}, the performance decline when removing part correlations in part representations demonstrate that, 
our proposed part correlations help in the geometric assembly of fractured parts,
as it is significant to aggregate geometric information between parts for geometric shape assembly.

As shown in Table~\ref{table:NSMab} and~\ref{table:BBab},
the translation embedding and adversarial training help improve the performance of our method, as described in Section~\ref{sec:trans} and ~\ref{sec:adv}, translation embedding and adversarial learning and can serve as pose fine-tuners and improve pose predictions. 



\section{Conclusion}

In this paper,
to tackle 3D geometric shape assembly tasks that rely on \textit{geometric} information of fractured parts,
we propose to leverage SE(3)-equivariant representations that disentangle shapes and poses to facilitate the task.
Our method leverages SE(3) equivariance in part representations considering part correlations, by learning both SE(3)-equivariant and -invariant part representations and aggregating them into SE(3)-equivariant representations.
To the best of our knowledge,
we are the first to explore leveraging SE(3) equivariance on multiple objects in related fields.
Experiments demonstrate 
the effectiveness of our method.

\vspace{-5mm}

\paragraph{Limitations \& Future Work}
In Breaking Bad, although we perform better than all baselines, this does not mean that we have solved the problem.
When the number of
fractures increases, the problem's complexity increases sharply, and most existing methods cannot perform well.
To completely solve the problem, more additional designs need to be added, while leveraging SE(3) equivariance is orthogonal to many designs.
For the whole framework, while the learned representations are equivariant to input part poses, the rotation regressor is non-equivariant, as it limits the degree-of-freedom in pose prediction and leads to worse results.
Besides,
it will take computing resources and time to train equivariant networks than ordinary networks. 

\vspace{-2mm}


\section{Acknowledge}
This work was supported by National Natural Science Foundation of China (No. 62136001).

{\small
\bibliographystyle{ieee_fullname}
\bibliography{egbib}
}

\end{document}